# Vers un modèle du « toucher vocal » pour la communication ubiquïte

Ambre Davat[1,2]  Véronique Aubergé[2]  Gang Feng[1]
(1) Univ. Grenoble Alpes, Grenoble INP, GIPSA-lab, F-38040 Grenoble, France
(2) CNRS, Grenoble INP, LIG, 38000 Grenoble, France
ambre.davat@gipsa-lab.grenoble-inp.fr, veronique.auberge@univ-grenoble-alpes.fr, gang.feng@gipsa-lab.grenoble-inp.fr

## Resume

Un des enjeux de la robotique de téléprésence est d'offrir une immersion socio-relationnelle ubiquïte. Pour y parvenir, il est nécessaire de comprendre et modéliser les facteurs qui permettraient au téléopérateur de contrôler la transmission de ses productions vocales, afin de lui en donner la perception, la proprioception et l'inter-proprioception. Ce modèle de transfert de la distance vocale socialement incarnée devra tenir compte de l'ensemble des paramètres impliqués dans l'effet socio-relationnel des productions vocales, en particulier l'intensité. Il devra également intégrer des éléments de contexte pertinents à la performance intentionnelle du locuteur téléopérant. Nous présentons dans cet article une première expérience visant à mesurer, analyser et modéliser la manière dont l'humain perçoit la distance physique qui le sépare d'un interlocuteur, en fonction de variations socio-affectives dans les productions vocales de cet interlocuteur. Ces résultats seront la référence des modèles qui seront implantés sur notre robot de téléprésence : RobAir Social Touch.

## Abstract

**Towards a model of "social touch" for ubiquitous communication**

One of the challenges of telepresence robotics is to provide ubiquitous social-interpersonal immersion. In order to achieve this, there is a need to understand and model the factors that would allow the users to control the transmission of their vocal productions, and to give them perception, proprioception and inter-proprioception of this control. This model for transferring socially embodied vocal distance should take into account all parameters involved in the social-interpersonal effect of vocal productions, especially intensity. It should also integrate the background information which is relevant for the speakers to express their intentions. In this paper, we present a first experiment for measuring, analyzing and modeling how human beings perceive the distance to an interlocutor, depending on socio-affective variations in the vocal productions of this interlocutor. These results will be the reference for the models which will be implanted in our telepresence robot: Robair Social Touch.

# 1 Introduction

Après l'invention de la téléphonie, qui permet de transporter la voix de quelqu'un d'un endroit à un autre, puis celle de la visiophonie, qui transmet également l'image, la robotique de téléprésence constitue une nouvelle étape dans l'immersion ubiquïte. En effet, il ne s'agit plus simplement de transporter la voix et l'image d'une personne vers un point de l'espace fixé par ses interlocuteurs, mais de permettre à l'utilisateur de contrôler ce point et de le déplacer dans l'espace distant dans lequel il communique à travers le « corps » physique d'un artefact robotique.

Ces robots de téléprésence reposent sous un nouvel angle la question de la fidélité avec laquelle la parole est transmise grâce aux moyens de télécommunication, question qui ne se pose plus pour le téléphone ou la visiophonie car nous avons culturellement intégré dans nos usages leurs artefacts. Ainsi par exemple, lors d'une communication téléphonique classique, les participants ont l'habitude de régler le volume sonore de leur interlocuteur au niveau qu'ils jugent acceptable. Une personne qui utilise un téléphone n'a donc pas à se préoccuper de savoir si sa voix est forte ou faible dans l'environnement distant car ce contrôle est clairement de la responsabilité de l'interlocuteur. En revanche, le pilote d'un robot de téléprésence peut être considéré par ses interlocuteurs comme maître de ce contrôle et donc responsable des artefacts liés à la télécommunication, d'autant plus que ses interlocuteurs sont réticents à l'idée de modifier les réglages du robot, car il est perçu non plus comme un simple objet de télécommunication, mais bien comme une personne physiquement présente (Paepcke et al., 2011). De plus, l'usage des robots de téléprésence n'est pas limité à une communication ponctuelle et limitée dans le temps : le téléopérateur doit pouvoir aller d'une pièce à une autre, initier une conversation, la conclure, ou encore se glisser au sein d'un groupe en pleine discussion, de la même manière qu'il le ferait s'il était présent en chair et en os. La téléprésence robotique nécessite ainsi certainement un contrôle fin de la portée vocale du téléopérateur, qui lui permette d'être audible de ses interlocuteurs, tout en respectant les règles de politesse en vigueur dans le lieu où se déroule l'interaction. Quelques systèmes ont déjà été proposés pour résoudre ce problème. Ils peuvent être classés en trois catégories :

- contrôle automatique : un algorithme permet d'adapter le volume audio en sortie du robot en fonction du contexte (Takahashi et al., 2015) ;
- sidetones insitatifs : le retour que le téléopérateur a de sa propre voix est amplifié artificiellement ; en conséquence, celui-ci diminue inconsciemment son intensité vocale : c'est ce qu'on appelle le « sidetone amplification effect » (Paepcke et al., 2011) ;
- interface visuelle : il s'agit de fournir au téléopérateur les informations nécessaires pour qu'il puisse contrôler de lui-même l'intensité de sa voix (Kimura et al. 2007).

Parmi ces trois grandes familles de solutions, seule la dernière permet au téléopérateur de contrôler lui-même la manière dont sa voix est transmise par le robot. C'est donc ce type d'interface que nous souhaitons développer. Pour cela, nous avons besoin d'un modèle de la portée vocale, c'est-à-dire un modèle qui lie le signal émis à la distance de communication. La recherche d'un tel modèle nous a conduit à développer un protocole expérimental afin d'étudier la manière dont la perception acoustique de l'espace peut être influencée socialement. Nous expliciterons d'abord le problème que nous nous posons, ensuite nous présenterons le protocole d'expérimentation que nous avons conçu puis mis en œuvre, et dont la complexité est relative à la complexité du problème posé. Enfin nous présenterons nos tous premiers résultats, sur seulement trois sujets, mais qui permettent déjà d'observer des tendances permettant de valider la pertinence du problème tel que nous le posons.

## 2   Toucher vocal et communication ubiquïte

Si les utilisateurs de robots de téléprésence n'arrivent pas à « toucher » acoustiquement leurs interlocuteurs comme ils le souhaitent, c'est parce qu'ils se trouvent dans deux environnements différents. En effet, comme les productions vocales dépendent de multiples éléments de contexte (Cooke, 2014), le signal qui serait adapté dans l'environnement du téléopérateur, ne l'est pas forcément dans celui où se trouve le robot. Par exemple, si le téléopérateur se trouve dans une pièce calme, sa voix transmise par le robot ne sera peut-être pas suffisamment forte pour être audible dans un environnement bruyant. Il est donc intéressant de pouvoir l'amplifier artificiellement, en ayant conscience toutefois que cette correction peut engendrer des artefacts sonores, des voix « schizophoniques » qui ne pourraient pas être produites naturellement (Schafer, 1977).

Cette schizophonie a sans doute un impact sur l'effet socio-relationnel des signaux vocaux. En effet, elle est notamment utilisée par la publicité, le cinéma et la musique afin de suggérer une certaine distance sociale entre l'auditeur et le locuteur (Maasø, 2008). Par exemple, un enregistrement qui parvient à capturer les bruits de bouche et la respiration du locuteur suggère une forme d'intimité à l'auditeur, qui ne devrait pas être capable de percevoir ces sons sans qu'ils soient extrêmement proches physiquement (Collins, Dockwray, 2015). Par ailleurs, (Gardner, 1969), (Brungart, Scott, 2001) ou encore (Philbeck, Mershon, 2002) ont montré que la manière dont un sujet perçoit à l'aveugle la distance qui le sépare d'une source de parole dépend de l'effort vocal utilisé au moment de la production du stimulus. Ainsi, un chuchotement est systématiquement perçu plus proche qu'il ne l'est réellement, tandis qu'un cri est perçu comme plus éloigné. La réponse des sujets est donc influencée par la distance de communication, et non simplement par la distance physique des sources sonores.

Un modèle du toucher vocal devrait donc tenir compte non seulement de nombreux éléments de contexte, mais également de la nature du signal produit par le locuteur. Partant de ce constat, nous avons voulu étudier si des variations socio-affectives portées par la parole, et plus précisément par la prosodie, peuvent influencer la manière dont un sujet perçoit acoustiquement la position spatiale de son interlocuteur. Pour décrire cette position spatiale perçue, nous nous intéressons à trois facteurs primitifs de la proxémie, potentiellement utilisés comme vecteurs socio-affectifs : la distance entre le sujet et son interlocuteur, mais aussi la direction dans laquelle se trouve l'interlocuteur, et sa posture regardant ou tournant le dos au sujet. L'objectif final de cette étude est de parvenir à modéliser la distance sociale perçue à partir de ces trois facteurs.

## 3   Protocole expérimental

L'expérience se déroule sur la plateforme d'expérimentation Domus, du LIG. Il s'agit d'une salle de 7.1 x 8.6 m, que nous avons aménagé avec des paravents pour former un espace carré de 7.1 x 7.1 m, soit 10 m en diagonale. Nous avons mesuré un temps de réverbération de 0.8 s[1]. C'est donc une salle particulièrement réverbérante, dans laquelle une simple mesure de l'intensité acoustique ne permet pas de deviner la distance d'une source sonore. Une photo du dispositif expérimental apparaît en Figure 1.

Notre protocole s'inspire initialement des tests psychoacoustiques classiques, lors desquels un sujet doit estimer à l'aveugle comment la source sonore qu'il entend est positionnée dans l'espace.

---

[1] Il s'agit du temps nécessaire pour que l'intensité du son diminue de 60 dB.

Cependant, afin de mettre en évidence un effet de variables socio-affectives sur la perception de l'auditeur, il est fondamental que le sujet ne soit pas conscient que nous mesurons sa capacité à repérer l'espace physique en fonction de variations de la distance sociale des stimuli, car alors le sujet procéderait à des méta-traitements cognitifs qui ne sont pas ceux que nous souhaitons mesurer directement. Pour assurer cette démarche d'observation en situation écologique, nous avons dû monter un scénario de type « caméra cachée » et baser notre expérience sur une tâche prétexte. De plus, comme ce sont les performances de régulation sociale humaine qui nous intéressent, nous avons décidé, sur ces deux contraintes, d'accepter les variabilités de production des stimuli inhérentes à l'écologie naturelle de production des humains interactants. Ainsi, dans cette expérience, la source sonore n'est pas un haut-parleur, mais un locuteur expert, dont les productions vocales seront analysées a posteriori pour contrôler leur régularité.

## 3.1 Scénario prétexte

Les sujets sont recrutés pour passer une expérience sur la perception du goût et de l'odorat. Cette expérience est censée faire partie d'un projet franco-japonais qui s'intéresse à la dimension culturelle et sociale des saveurs et des odeurs. Elle se déroule en binôme : un des sujets doit goûter des mini-pilules gustatives, l'autre respirer des boîtes à odeurs. Avant l'expérience, ils doivent remplir un questionnaire concernant leur pratique gustative et olfactive, et qui permet de recueillir des informations sur leur accent. Au moment où le sujet (S) se présente pour passer l'expérience, il rencontre le locuteur expert (L), présenté comme le second sujet de l'expérience. (L) se fait passer pour un nez professionnel, ce qui justifie qu'il puisse produire des énoncés à la fois très autoritaires et très hésitants. Un expérimentateur (E) présente l'expérience, en s'appuyant sur les questions posées par (L) afin de convaincre (S) qu'il passe bien une expérience sur le goût et l'odorat.

(S) comprend ainsi que chaque mini-pilule est assortie à une boîte à odeur. A chaque étape, (L) goûtera une mini-pilule disposée dans des gobelets répartis sur deux rangées de tables et annoncera ce qu'il a reconnu. Ensuite, (E) donnera une boîte à odeur à respirer à (S), qui devra alors discuter avec (L) jusqu'à ce que chacun donne un avis définitif. Les deux participants seront placés dans différentes situations d'interaction, soit disant pour observer des variations de leurs capacités perceptives. Afin d'éviter qu'ils puissent lire sur le visage de l'autre des informations de plaisir ou de déplaisir qui pourraient influencer leur perception, (S) devra porter un masque qui l'aveugle et dissimule le bas de son visage. Il passera donc l'expérience assis sur une chaise au centre de la pièce, à côté de (E). Cependant, cette situation serait très inconfortable pour (L), qui aurait l'impression de parler à quelqu'un qui l'ignore. Ainsi, pour s'assurer que (S) reste concentré sur sa tâche et rassurer (L), on demande à (S) d'indiquer avant chaque boîte à odeur :

- la direction dans laquelle se trouve (L) (avant, gauche, derrière ou droite)
- sa distance (devant la première table, entre les deux tables ou derrière la deuxième table)
- son orientation (face au sujet, ou dos au sujet)

Par ailleurs, (L) reçoit un ordre de passage indiquant les positions des gobelets et leur numéro. Au dos de cette feuille se trouve les indications pour la vraie expérience, à savoir :

- la direction (avant, gauche, derrière ou droite)
- la distance (proche, milieu ou loin)
- l'orientation (face au sujet, ou dos au sujet)
- l'intensité à produire (faible ou forte)
- le socio-affect à communiquer (confiance autoritaire ou doute poli)
- le mot-clé à prononcer (ex : pomme, orange...)

Entre chaque étape, de la musique est diffusée par des hauts parleurs placés derrière les oreilles du sujet pour le faire patienter et dissimuler les pas du locuteur expert, lequel a prétexté dès le début de l'expérience avoir une fracture de l'orteil pour pouvoir ôter ses chaussures et se déplacer en chaussettes. Le sujet et le locuteur expert portent un micro serre-tête Sennheiser HSP 4 relié à un émetteur radio afin que leur échange soit enregistré.

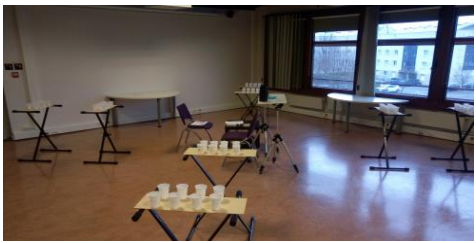

FIGURE 1 : Photo du dispositif expérimental

A la fin de l'expérience, un débriefing est effectué. Il permet d'abord de vérifier que le sujet n'a pas deviné le but réel de l'expérience. Ensuite, la supercherie est dévoilée, et on s'assure que le sujet a bien compris les buts de l'expérience pour qu'il puisse donner son consentement éclairé.

Il s'agit d'une expérience lourde à monter, à mettre au point, puis à reproduire pour chaque sujet, puisqu'elle dure environ 1h30, introduction et débriefing compris. Il est donc important de noter que les 3 sujets déjà enregistrés, ainsi que plusieurs sujets préalables, non conservés tant que la mise au point n'était pas stabilisée, n'ont pas montré de signe d'ennui, de désintérêt ou de charge cognitive trop forte par rapport à la tâche prétexte ; en outre, nous n'avons observé a priori ni d'effet d'apprentissage, ni de dégradations des réponses des sujets au cours de l'expérience (nous le vérifierons statistiquement quand nous aurons plus de sujets). Nous allons à présent détailler la manière dont ce dispositif expérimental a été choisi.

## 3.2    Choix des distances

Pour définir les distances mises en jeu dans cette expérience, nous nous sommes intéressés à l'incertitude avec laquelle un sujet estime la distance d'une source sonore. En effet, la psychoacoustique a montré que lorsqu'on demande à un sujet d'estimer à plusieurs reprises la distance d'une même source sonore, ses réponses varient. En choisissant des distances suffisamment proches les unes des autres, il est donc possible d'induire le sujet en erreur. Au contraire, si les distances sont trop différentes, le locuteur n'a aucune difficulté à deviner si son interlocuteur est devant, derrière, ou entre les deux tables. Cette incertitude dépend probablement des caractéristiques acoustiques de la salle où se déroule l'expérience, et varie d'un sujet à l'autre. Ainsi, (Zahorik et al., 2005) évoquent un flou perceptif évalué entre 5 et 20 % de la distance effective d'après un article de Haustein de 1969, et entre 20 et 60 % lorsque les distances sont représentées en échelle logarithmique d'après réanalyse de leurs propres travaux. Dans un autre article, (Calcagno, Abregú, 2012) indiquent l'écart type des distances estimées par leurs sujets. Celui-ci augmente linéairement dans le cas où les sujets doivent estimer la distance des sources sonores à l'aveugle. A titre indicatif, l'écart type est d'environ 40 cm pour une source située à 2 m, lorsque les sujets ont eu l'occasion de voir la salle avant le début du test. Par ailleurs, (Anderson, Zahorik, 2014) montrent que l'erreur de jugement des sujets suit une distribution normale lorsque la distance est représentée en échelle logarithmique.

Après un premier test avec des tables de 80 x 160 cm de large, nous avons fabriqué des tables plus petites en utilisant des panneaux de bois de 20 x 60 cm posés sur des trépieds. Finalement, nous avons choisi de placer deux rangées de tables, respectivement à 2 et 3 m de l'emplacement du sujet. Le locuteur se place donc à 1 m 70, 2 m 50 ou 3 m 30 du sujet. Non seulement cet aménagement convient parfaitement aux dimensions de la pièce, mais il est intéressant en termes perceptifs. D'une part, il permet d'étudier le mode proche et le mode éloigné de la sphère sociale définie selon la théorie proxémique (Hall, 1966). D'autre part, les études psychoacoustiques ont montré que nous avons tendance à surestimer la distance des sources sonores proches et à sous-estimer celles des sources éloignées. La distance à laquelle s'inverse la tendance dépend des propriétés acoustiques de la pièce, mais semble être dans l'ordre de grandeur des distances que nous utilisons. Ainsi, (Anderson, Zahorik, 2014) ont observé un point d'inflexion à 1 m 90, puis à 3 m 22 dans une salle plus réverbérante, dans laquelle les distances sont perçues plus éloignées.

### 3.3   Choix des directions et orientations

Dans cette expérience, il ne s'agit pas de vérifier si nos sujets sont capables d'évaluer la direction d'arrivée des sons, mais de pouvoir éventuellement observer des biais perceptifs différents selon la manière dont le sujet et son interlocuteur sont positionnés dans l'espace. Parler en étant rigoureusement face à face avec quelqu'un, ce n'est pas la même chose que de lui parler de profile, ou même de dos. En effet, l'orientation du locuteur a d'abord une conséquence acoustique, car la voix humaine a une directivité : au lieu de se propager uniformément dans toutes les directions de l'espace, la puissance acoustique se répartie sous une forme cardioïde (Chu, 2002). En particulier, les hautes fréquences de la voix sont particulièrement atténuées derrière la tête du locuteur. En conséquence, un auditeur est capable de deviner à l'aveugle l'orientation de la tête d'un locuteur (Edlund et al., 2012). Par ailleurs, l'orientation du corps a un sens social ; elle est donc généralement prise en compte dans les études sur la proxémie, soit directement, par exemple comme dans (Remland et al.), soit indirectement, lorsque c'est le regard des interlocuteurs qui est analysé.

### 3.4   Choix des distances sociales des productions vocales

Le locuteur expert doit être capable d'exprimer deux socio-affects différents : une confiance autoritaire par laquelle il marque une distance sociale grande et de dominance avec son interlocuteur, et un doute poli, par lequel il se rapproche socialement de son interlocuteur et inverse la dominance. Intrinsèquement à la nature prosodique de ces énoncés, le locuteur a tendance à parler plus fort pour exprimer la confiance autoritaire, et plus doucement pour exprimer le doute. Nous avons donc ajouté pour le locuteur complice une consigne de contrôle de son intensité de production, faible ou forte, sur chacune des deux variables socio-affectives, afin d'observer si celle-ci a un impact sur la perception de l'auditeur.

## 4   Analyse des premiers résultats

Notre objectif est de faire passer entre 20 et 30 sujets sur cette expérience. Nous présentons ici les résultats obtenus avec nos 3 premiers sujets.

## 4.1 Vérification de la régularité de production du locuteur expert

Il est important de vérifier que notre locuteur expert arrive à produire les différents socio-affects et intensités demandés, tant au niveau des contenus attitudinaux que des réalisations acoustiques. Pour mesurer l'intensité, nous avons choisi la procédure suivante, réalisée par un script Praat :

1. extraction du pitch et de l'intensité du mot-clé (il s'agit bien sûre de l'intensité de l'enveloppe, et pas de l'intensité instantanée)
2. échantillonnage toutes les 10ms du pitch et de l'intensité
3. moyennage des intensités pour lesquels le pitch est défini

Les intensités mesurées à partir des enregistrements du micro porté par le locuteur expert sont présentées dans le Tableau 1. En moyenne, il y a bien un écart de plus de 7 dB entre les mots qui devaient être prononcés avec une intensité faible, et ceux devant être prononcés avec une intensité forte. Même si l'écart type ainsi que les valeurs min et max indiquent que l'intensité de certains mots-clés ne convient pas à leur catégorie, le locuteur expert arrive donc le plus souvent à suivre les consignes concernant les variations d'intensité. Cette méthode de mesure est néanmoins sensible à la position du micro, qui varie d'une expérience à l'autre, voire au cours d'une même expérience. En effet, ce micro n'est situé qu'à 2-3 cm de la bouche du locuteur, donc la moindre variation de son écartement fait varier l'intensité mesurée. Pour avoir une mesure vraiment fiable, il faut utiliser un autre micro, par exemple celui porté par le sujet. Cependant, si celui-ci fait le moindre geste pendant que le mot-clé est prononcé, le bruit engendré fausse la mesure d'intensité. A l'avenir, nous utiliserons donc un micro supplémentaire, placé à la verticale du sujet.

| Attitude | Doute poli | | Confiance autoritaire | |
|---|---|---|---|---|
| Intensité | Faible | Forte | Faible | Forte |
| Min | 43,9 | 50,7 | 45,9 | 56,3 |
| Max | 62 | 70,6 | 65,1 | 69,9 |
| Moyenne | 53,8 | 61,9 | 55,3 | 62,6 |
| Ecart type | 3,2 | 3,1 | 3,9 | 2,5 |

TABLEAU 1 : Mesure de l'intensité (dB) des mots-clés prononcés par le locuteur expert

Un test perceptif devra également être effectué pour confirmer que les attitudes sont bien reconnues, quelle que soit l'intensité utilisée. Nous n'avons pas encore fait d'analyse rigoureuse des différents mots-clés en terme de qualité de voix et de prosodie, mais nous avons déjà pu remarquer qu'il est difficile pour le locuteur expert d'exprimer une confiance autoritaire d'une voix faible et un doute poli d'une voix forte. A l'oreille, la stratégie qui semble efficace consiste dans un cas à parler très vite pour avoir une voix sèche, et dans l'autre à faire trainer les mots-clés et leur donner une tournure interrogative.

## 4.2 Premières observations

Une première manière d'analyser les réponses données par les sujets est d'étudier leur répartition en fonction des cinq variables étudiées. Par manque d'espace, nous ne représenterons pas les résultats liés à la distance, car outre les confusions avant/arrière régulièrement observées en psychoacoustique, les sujets se trompent rarement pour estimer la distance. Pour simplifier la mise en page, les quatre variables et les deux facteurs proxémiques retenus sont présentés au même niveau, mais bien évidemment, ils n'interviennent pas de la même manière dans la distance sociale ressentie par les sujets. Une lecture rapide de la Figure 2 permet d'observer une corrélation entre la position réelle du locuteur et celle perçue par les sujets, et on peut d'ores et déjà noter quelques tendances intéressantes : 1) les sujets n'ont pas de difficulté à percevoir la distance proche, en revanche, ils confondent les deux autres distances ; 2) les sujets sont indécis lorsqu'ils doivent estimer l'orientation, pourtant ils ne se trompent quasiment jamais lorsque le locuteur est de dos ; 3) lorsque le locuteur est de dos ou 4) son intensité faible, la distance perçue par le sujet augmente.

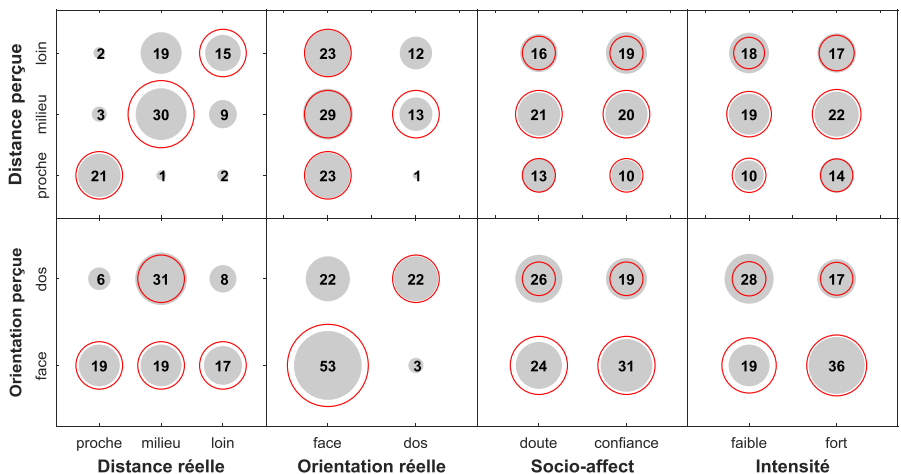

FIGURE 2 : Décomptes des réponses données par les sujets en %
*les cercles rouges correspondent au décompte qu'on aurait obtenu s'ils avaient répondu juste*

## 5 Conclusion

L'interaction face à face in situ est un système complexe et fin qui permet aux interactants de se situer autant dans l'espace physique que dans l'espace social, la proxémie de leurs déplacements relatifs utilisant, différemment selon leurs cultures, cet espace physique pour signifier des informations sur leur espace social. Par cette étude, nous voulons montrer qu'il en est de même dans l'espace acoustique et que nous intégrons les distances physiques à nos productions vocales et notre audition pour exprimer et percevoir les distances socio-relationnelles. Ainsi, pour pouvoir assurer une immersion ubiquïte aussi bien physique que sociale, il faudra assister le téléopérateur dans la gestion de ces contrôles fins sous peine de générer des malentendus socio-relationnels importants. Nos premiers résultats sont plutôt encourageants. S'ils se confirment, nous pourrons en déduire un modèle du toucher vocal, qui sera testé et raffiné grâce à l'implémentation d'une interface pour robot de téléprésence.

## Références